%% file: main.tex
\documentclass[letterpaper, 10 pt, conference]{ieeeconf}  % Comment this line out if you need a4paper

\IEEEoverridecommandlockouts        % This command is only needed for thanks command
\input{sections/0-preamble}
\title{\LARGE \bf
Minimal Work: A Grasp Quality Metric for Deformable Hollow Objects
}

\author{Jingyi Xu$^{1,2}$, Michael Danielczuk$^1$, Jeff Ichnowski$^1$, Jeffrey Mahler$^1$, Eckehard Steinbach$^2$, Ken Goldberg$^1$
\thanks{$^{1}$The AUTOLAB at UC Berkeley, CA, USA. $^{2}$Technical University of Munich, Chair of Media Technology, Germany.  Email: \{ jingyi\_xu, mdanielczuk, jeffi, jmahler, goldberg\}@berkeley.edu, eckehard.steinbach@tum.de}%
}

\makeatletter
\def\endthebibliography{%
	\def\@noitemerr{\@latex@warning{Empty `thebibliography' environment}}%
	\endlist
}
\makeatother

\begin{document}
\maketitle 
\thispagestyle{empty}
\pagestyle{empty}

%%%%%%%%%%%%%%%%%%%%%%%%%%%%%%%%%%%%%%%%%%%%%%%%%%%%%%%%%%%%%%%%%%%%%%%%%%s%%%%%%

\input{sections/1-abstract}

% For peer review papers, you can put extra information on the cover
% page as needed:
% \ifCLASSOPTIONpeerreview
% \begin{center} \bfseries EDICS Category: 3-BBND \end{center}
% \fi
%
% For peerreview papers, this IEEEtran command inserts a page break and
% creates the second title. It will be ignored for other modes.

\section{Introduction}
\input{sections/2-introduction}

\section{Related Work}
\input{sections/3-related_work}

\section{Problem Statement}
\input{sections/4-problem-statement.tex}

\section{Minimal Work Grasp Quality Metric}
\input{sections/5-methodology}

% \input{sections/6-methodology-2}

\section{Algorithm}
\input{sections/6-implementation.tex}

\section{Experiments and Results}
\input{sections/7-experiments}

\section{Discussion and future work}
\input{sections/8-discussion}

% \section{Discussion and future work}
% \input{sections/9-conclusion}

\input{sections/10-ackowledgement.tex}

\bibliographystyle{IEEEtran}
\bibliography{IEEEfull,main}

\end{document}

%% file: sections/1-abstract.tex
\begin{abstract}
Robot grasping of deformable hollow objects such as plastic bottles and cups is challenging as the grasp should resist disturbances while minimally deforming the object so as not to damage it or dislodge liquids.
%To compute optimal grasp placement, standard wrench-based quality metrics only consider the ability of a grasp to resist external forces and torques, without considering potential deflection of the object surface.
% Deformable jaws or gripper with compliant materials are widely used to increase grasp stability and prevent object damage. 
We propose minimal work as a novel grasp quality metric that combines wrench resistance and the object deformation. %by formulating grasp planning as an optimization problem.   
%A task is formulated as a 6D target wrench to be counterbalanced by the grasp. %or a 6D unit sphere if unknown.
We introduce an efficient algorithm to compute required work to resist an external wrench for a manipulation task by solving a linear program. 
% We measure the deformability of objects at different locations with physical experiments. 
The algorithm first computes the minimum required grasp force and an estimation of the gripper jaw displacements based on the object's deformability at different locations measured with physical experiments. 
% The object deformation is then computed based on the grasp force and the deformability of the object.
The work done by the jaws is the product of the grasp force and the displacements. 
% Grasps requires minimal work have a high quality. 
The grasp quality metric is computed based on the required work under perturbations of grasp poses to address uncertainties in actuation. 
We collect 460 physical grasps with a UR5 robot and a Robotiq gripper.
Physical experiments suggest the minimal work quality metric reaches 74.2\% balanced accuracy  and is up to 24.2\% higher than classical wrench-based quality metrics, where the balanced accuracy is the raw accuracy normalized by the number of successful and failed real-world grasps. 
\end{abstract}

%% file: sections/2-introduction.tex
% 1. motivation
%     why grasping important
%     why wrench resistance important
    
% 2. how state of the art is solving the problem
%     minimal force grasps
%     planar object, jia, rotate rob
%     3D object but no quantitative quality
%     can not adapt to different tasks 
% 3. how we tackle the problem
%     why using REACH
%     collect stiffness
% 4. how do we demonstrate the results 

% minimal force highly depend on the task. 
% if object has different stiffness, minimal force not enough 

% Robot grasping is an active research area in both industrial and household environments.
For rigid objects, wrench-based quality metrics~\cite{ferrari.1992,li1988task} are widely used to optimize grasp placements and to estimate grasp success~\cite{pollard1994parallel,jameson1985analytic} since they quantify grasps and are suitable for both general and task-oriented grasps.

Grasping deformable objects is more challenging.
In addition to resisting external disturbances, grasps should minimize the deformation of the object to avoid damage or dislodging liquids e.g. when grasping plastic cups and bottles. 
% Fig. \ref{fig:headshot} shows an example of grasping a plastic bottle without considering the object deformation. 
% Minimum force grasps \cite{howard2000intelligent,delgado2015tactile} are typically used to grasp such objects to minimize object deformations. 
% Jia~et~al.~\cite{jia2014grasping} propose minimum work to hold deformable planar objects, while Ramirez-Alpizar~et~al.~\cite{ramirez2012dynamic} focus on  rotating planar objects, which are placed on a plate. 
% Other works use the Finite Element Method (FEM) to estimate object deformation and lift up objects such as tomatoes and foams by checking stick-slip constraints constantly to consider the object's new geometry \cite{lin2015picking,zaidi2017model}.
Existing grasp planning for deformable objects focus either on holding deformable planar objects~\cite{gopalakrishnan2005d,jia2014grasping,ramirez2012dynamic} or lifting 3D objects with a pre-selected grasp placement~\cite{lin2015picking,zaidi2017model}. 
%but the former does not consider the gravity, while the latter does not quantify a grasp and are, therefore, not applicable to plan grasps for 3D objects.
%Furthermore, we should be able to plan grasps for more sophisticated manipulation tasks for both rigid and deformable objects, such as lifting and rotating a mug or a plastic cup for pouring. 

\begin{figure}[t!]
	\centering	
    \includegraphics[width=90 mm,trim={0cm  0cm 0cm 0cm},clip]{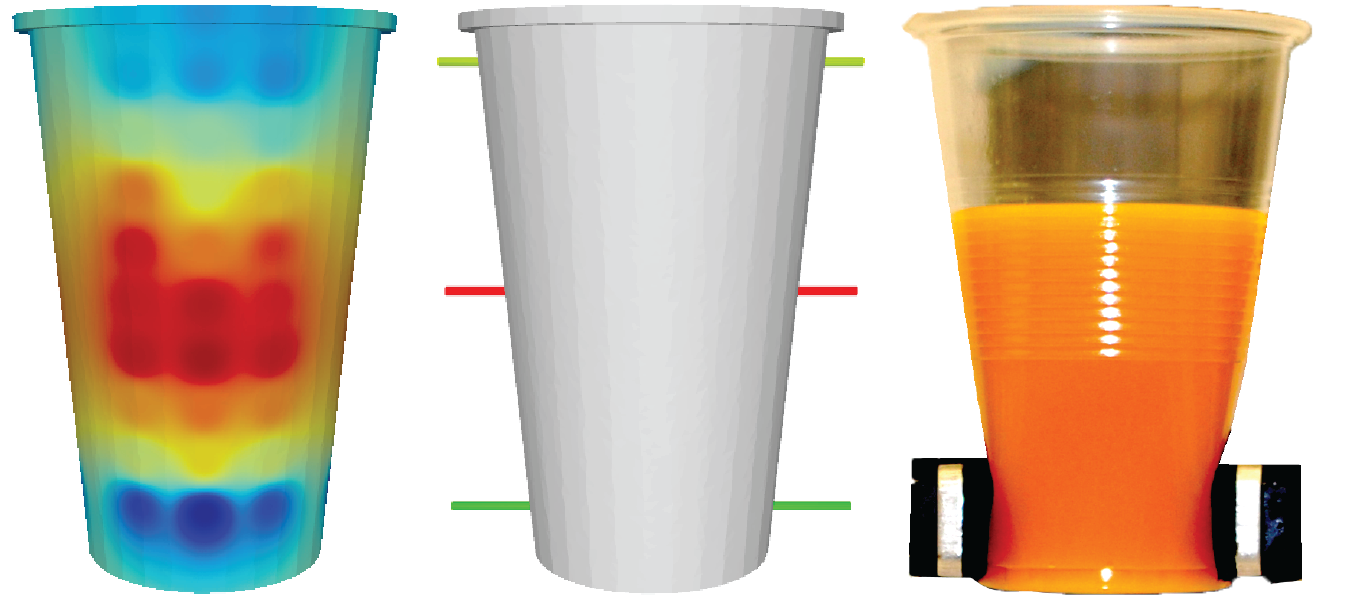}
	\caption{Plastic cup example. Left: stiffness of cup, red indicates a low stiffness. Middle: three planned grasps, red indicates high work. Right: minimal work grasp in physical experiment. }
	\label{fig:headshot}
\end{figure}

We propose the \emph{minimal work} quality metric,
a novel quality metric that considers both wrench resistance and object deformation.
Fig. \ref{fig:headshot} shows an example of planned grasps for a plastic cup with the proposed metric.
% a novel quality metric that unifies grasp quality for deformable gripper pads for both rigid and 3D deformable hollow objects.
% The proposed metric is applicable to grasping deformable hollow objects with rigid jaws. 
% The proposed metric resists external disturbance wrenches while minimizing object deformation.
To compute the quality of a grasp, we first estimate the minimum required grasp force to resist an external wrench with the un-deformed object's shape.
We use the Robust Efficient Area Contact Hypothesis (REACH) model~\cite{danielczuk2019reach} to estimate the contact area and the pressure distribution by means of the constructive solid geometry intersection of the extruded polygon of the jaw with the object.
The friction wrenches of the non-planar area contacts are then modeled with a 6D ellipsoidal limit surface \cite{xu2019tro}.
We formulate an optimization problem to solve the minimum grasp force subject to the friction constraints. %With the minimum required grasp force, we then estimate the object deformation %based on the object stiffness at the grasp location. 
%The stiffness is experimentally determined by squeezing the object at different locations with a known force and collect the gripper opening.
The required work of the jaws to complete the task provides the metric.
The object's deformation is considered as an approximation of the jaw displacement, which are estimated from the minimum grasp force. 
We decouple the wrench analysis and the computation for object's deformation, so that the grasp quality can be efficiently computed without Finite Element Method (FEM) simulation for each grasp or repeatedly determining wrench resistance throughout the grasp. 
% We compare the proposed minimal work quality metric with wrench resistance and minimal force grasps for xxx objects. 

% Minimal force grasps are both time and energy efficient, especially for fragile objects or plastic containers filled with liquid. 
% The grasp should be stable, while the object has minimal deformation such that it will not be damaged or content remains in the container. 
% To select the optimal grasp placement, standard wrench-based quality metrics only consider the stability of a grasp [][][].
% % Wrench-based metrics typically choose a grasp with maximal volume or maximal shortest distance between the origin and all facets of the grasp wrench space (GWS). 
% It is non-trivial to find a stable grasp while minimizing the deformation. 

% We propose the \emph{minimal work} grasp quality metric, a novel metric that unifies grasp quality for both rigid and deformable objects. 
% We consider both grasp stability and object deformation by formulating the grasp placement optimization as a min-max problem. 
% The metric calculates the required work done by the grasp jaws for a manipulation task. A task is formulated as a 6D target wrench to be counterbalanced by the grasp or a a 6D unit sphere if unknown. 
% The grasp that requires minimal work for the task is considered optimal.

This paper provides the following contributions:
\begin{enumerate}
    \item A novel minimal work grasp quality metric for 3D deformable hollow objects that considers both wrench resistance and object deformation.
    \item An efficient algorithm to compute minimal required work to resist a 6D external wrench by solving a linear program.
    \item Physical experiments that suggest grasps planned with the minimal work quality metric lead to 74.2\% balanced accuracy, 24.2\% and 12.7 \% higher than the grasp reliability metric and the minimal force metric, respectively. 
\end{enumerate}

%% file: sections/3-related_work.tex
% We summarize related work of wrench-based grasp quality metrics in three aspects: grasps for rigid objects, task-oriented grasps, and grasps for deformable objects.
We summarize related work in wrench-based grasp quality metrics for rigid objects and grasp planning for deformable objects.
Excellent surveys for contact modeling can be found in \cite{rimon_burdick_2019,siciliano2016springer,kao2008contact,bicchi2000robotic}, for grasp quality metrics in \cite{roa2015grasp}, and for deformable object manipulation in \cite{sanchez2018robotic}.

% There is a substantial literature on contact modeling for robot grasping, with notable surveys by Kao, Lynch, and Burdick~\cite{kao2008contact}, Rimon and Burdick~\cite{rimon_burdick_2019}, and Bicchi and Kumar~\cite{bicchi2000robotic}.

\subsection{Grasping rigid objects}
%- force closure
%- GWS, epsilon, volume
%- grasp quality surveys from Maximo Roa
%- dex-net 2.0, uses force closure
A common grasp quality metric, force closure \cite{nguyen1988constructing}, evaluates a grasp by whether it can resist any 6D disturbance wrenches with an arbitrarily large grasp force. 
The volume of a grasp wrench space (GWS) \cite{li1988task} and the $\epsilon$-metric \cite{ferrari.1992} are also widely used to quantify grasp quality.
While the volume reflects the quality of the entire grasp, the $\epsilon$-metric identifies the weakest point of a grasp, as $\epsilon$ is the shortest distance between the origin and any facet of the GWS.
The GWS used to evaluate grasps is typically constructed with a bounded sum-magnitude of grasp forces for computational efficiency.
Krug et al. \cite{krug2017grasp} suggest that such a construction is over-conservative for fully actuated grippers, as the force of each jaw is limited independently. 

%- ellipsoid task wrench space
%- OWS 
%- how to select TWS
%- maximal task coverage, Yu Sun
%- another Yu Sun paper after that
%- related work in my IROS
%- task-directed grasp stability assessment, Yasmine
%- dex-net 4.0

External disturbance wrenches can be estimated for specific tasks.
Task-oriented grasp quality metrics are, therefore, better suitable to estimate wrench resistance for such tasks.
% Objects are frequently grasped for a specific task. 
A task wrench space (TWS) describes expected disturbance wrenches during a manipulation task.
The TWS is typically modeled with possible wrenches that can be imposed on an object \cite{pollard1994parallel,borst2004grasp}, or an 6D ellipsoid \cite{li1988task}.
The quality of a grasp is the scale of the TWS, so that it just fits into the GWS. 
In addition, the quality can be measured with the minimal required force for a task \cite{howard2000intelligent} or minimal coefficient of friction \cite{jameson1985analytic}.  
Lin and Yu \cite{lin2015grasp} observe that some disturbance wrenches happen more often than others during a task execution and, therefore, select the grasp whose GWS covers most frequent disturbances. 
They further select the optimal grasp in \cite{lin2015task}, which minimizes the required motion effort of the end effector to fulfill a certain task. 

\subsection{Grasping deformable objects}
%- min work for planar objects, Yanbin Jia
%- deform closure
%- bounded force closure
Manipulating deformable objects is an active area, with applications such as food handling \cite{long2014force}, fabric manipulation \cite{li2015folding,seita2019bedmake}, and elastic rod manipulation \cite{bretl2014quasi}.
When a frictionless grasp immobilizes a rigid object, it is defined as form closure. 
Gopalakrishnan and Goldberg \cite{gopalakrishnan2005d} generalize this concept to holding deformable objects with frictionless contacts, where a grasp is defined as deform closure, when positive work is required to release the object. 
Wakamatsu~et~al.~\cite{wakamatsu1996static} introduce the bounded force closure metric, which guarantees a force closure grasp under a maximal allowable external force. 
Delgado~et~al.~\cite{delgado2015tactile} reduce object deformation for a holding task by computing the maximum allowed force to be exerted on an object. 
Jia~et~al.~\cite{jia2014grasping} propose a grasping strategy to squeeze holding deformable planar objects based on work performed by the jaws. 
When two jaws squeeze and immobilize an object, and a third jaw tries to break the grasp by pushing the object, the translations of the two pushing jaws that minimize the required work to balance the object is selected. 
Since the metric targets planar objects, the 3D geometry or the gravity is not considered and a point contact model is used for friction analysis. Lin~et~al.~\cite{lin2015picking,lin2014picking} address the problem of lifting a deformable object based on an object mesh model and jaw positions. 
An FEM formulation computes the object deformation based on the jaw displacements.
The object will be lifted if the majority of the contact points are sticking. 
% The object deformation is computed by an FEM formulation based on the jaw displacements.
% If most of the contact points are sticking, the object will be lifted.
Similarly, Zaidi~et~al.\cite{zaidi2017model} use FEM simulation to manipulate objects with large deformations, such as objects made of foam or rubber. 
Alt~et~al \cite{alt.2016} use FEM simulation and heuristics to plan grasps for deformable thin-walled objects. 

Inspired by \cite{gopalakrishnan2005d} and \cite{jia2014grasping}, the proposed minimal work quality metric optimizes grasp placements to manipulate 3D deformable hollow objects.
Furthermore, the metric is suitable for many tasks that can be modeled as target wrenches to be resisted.

%% file: sections/4-problem-statement.tex
\subsection{Overview}
We consider the problem of grasp planning for 3D deformable hollow objects with compliant jaw pads based on the ability of a grasp to resist target wrenches and the deformability of the object at the grasp location. %, and 3) the deformation of the gripper jaws. 
%The metric is applicable to grasping rigid objects. 

\subsection{Assumptions}
We make the following assumptions:
\begin{enumerate}
	\item The geometry and the deformability are known for the objects to be grasped.
	\item Quasi-static analysis and Coulomb friction with a known coefficient of friction.
	\item A linear elastic model of soft jaw pads and objects.
% 	\item The contact area is constant.
\end{enumerate}

\subsection{Notation}
\begin{itemize}
    \item $\boldw \in \mathbb{R}^6$: wrench of a contact, which is concatenated by a 3D force and a 3D torque. 
    \item $\mathcal{C}_i$: the constraint set that limits the maximum possible friction wrench and the wrench impressed by the normal pressure of the $i$-th contact. 
    \item $\work$: the work performed by the gripper jaws
    \item $\boldsymbol{t} \in \mathbb{R}^6$: a target wrench to be resisted with a grasp
    
%     \item $\f \in \mathbb{R}^6$: the frictional wrench
%     % \item $\hat{\f} \in \mathbb{R}^6$: the frictional wrench under a unit normal force
%     \item $\fperp \in \mathbb{R}^6$: the wrench impressed by the normal pressure, defined as the normal wrench
%     % \item $\hat{\fperp} \in \mathbb{R}^6$: the normal wrench under a unit normal force 
%     % \item $N$: number of gripper jaws
%     \item $K$: number of perturbations of grasp pose
%     \item $\boldw_i = \f_i + \fperp_i$, $i \in \{1 \ldots N \}$: the total wrench of a contact 
%     \item $w$: work performed by the jaws
%     % \item $\t \in \mathbb{R}^6$: the target wrench to be resisted
%     % \item $\matr{A} \in \mathbb{R}^{6 \times 6}$: the matrix to model the 6D ellipsoidal limit surface 
%     \item $\boldsymbol{p}^{6 \times M}$: $M$ sampled points on the 6D ellipsoid
%     \item $\boldsymbol{n}^{6 \times M}$: normals the ellipsoid at $\boldsymbol{p}$
%     % \item $\mathcal{F}$: the set of contact wrench constraints
%     % \item $F = \sum_{i=1}^N F_i$: the grasp force equals sum of the force of each jaw
%     % \item $\x=[\x_0^T,\ldots,\x_N^T], \x_i=[\boldw_i, F_i]^T \in \mathbb{R}^7$: the variable to be solved by the optimization 
%     % \item $d_i$: the displacement of the $i$-th jaw
%     % \item $d_j$: the depth of the compliant jaw pads
%     % \item $s_j$: the stiffness of the compliant jaw pads
%     % \item $s_o$: the stiffness of the object 
\end{itemize}

\subsection{Metrics}
\mysum{add description for grasp reliability}
We compare the proposed minimal work grasp quality metric with two widely used metrics.
Each grasp is computed under $K$ perturbations of grasp poses to address uncertainties in actuation. 
\begin{enumerate}
    \item Grasp reliability metric $q_r$: $r = 1$ if the grasp is able to resist the target wrench $\t \in \mathbb{R}^6$ without exceeding the maximum closing force and $r = 0$ otherwise: 
    \begin{equation*}
        q_r = \frac{1}{K}\sum_{k=1}^K r_k.
    \end{equation*}
    \item Minimal force metric $q_f$: The minimal required grasp force to resist $\t$: 
\begin{equation*}
    q_f = \frac{1}{K}\sum_{k=1}^K \left(1 - \frac{F_k}{F_{\max}}\right). 
    % q_f = \sum_{k=1}^K(1 - \frac{\sum_{i=1}^n F_i}{F_{\max}})/K. 
\end{equation*}
where $F_{\max}$ is the maximal grasp force used in the experiments.  
\end{enumerate}

\subsection{Task modeling}
Two manipulation tasks are considered: 1) vertical lifting and 2) lifting and $90^\circ$  rotation. 
We model each task with a 6D gravity wrench to be resisted under multiple object poses obtained by discretizing the trajectory of the task. 
A single pose is considered for the lifting task, since the gravity wrench remains unchanged during the manipulation, while three poses are considered for the lifting and rotation task. 
% For each grasp, the lowest score among all object poses for a task is selected as the grasp score. 

\subsection{Objective}
We use the balanced accuracy score to evaluate the prediction accuracy of each quality metric by comparing them with real-world grasps. 
The predicted grasp success is binary and is true if the metric is higher than a threshold.
The balanced accuracy is suitable for imbalanced datasets and is computed with the raw accuracy, where each sample is weighted with the inverse prevalence of its true class.

%% file: sections/5-methodology.tex
To evaluate a grasp candidate, we compute the minimal work of the gripper jaws required to complete a manipulation task. %, where a task is formulated as a 6D wrench. %(\textbf{6D SPHERE NOT INCLUDED FOR NOW})
We first model the frictional contacts and compute the minimal grasp force by formulating Equation~(\ref{eq:sol_min_force}) as a linear program (LP).
We then estimate the deformation based on the force and object stiffness at the contact locations. 
The work of each gripper jaw is the product of grasp force and jaw displacement.
The sum of the work of each jaw forms the work of the grasp. 

The proposed algorithm to compute the minimal work can use different contact models and object's stiffness acquisition methods.
We use the REACH model \cite{danielczuk2019reach} and a 6D ellipsoidal limit surface to describe a contact.
The stiffness is collected with physical robots. 
Details can be found in Sec. \ref{sec:implementation} and \ref{sec:stiffness_acquisition}.
% \subsection{Grasp force computation for a known task}
% Here we describe the computation of minimal required grasp force for two fingers.
% This can be easily generalized to more fingers. 
\mysum{compute grasp force}
% Denote $\matr{G} \in \mathbb{R}^{6 \times 6n}$ as the grasp matrix for $n$ contacts, $\f \in \F$ as the friction wrench, $\fperp$ as the wrench impressed by the normal force. 

% We set the variable of the LP to be $\x = [\x_0^T,\ldots,\x_n^T]^T$, where $\x_i = [\boldw_i^T, F_i]^T$ and $F_i$ is the magnitude of the grasp force.
For a grasp with $N$ contacts, denote $\matr{G} \in \mathbb{R}^{6 \times 6N}$ as the grasp matrix, $\boldsymbol{w}_i \in \mathbb{R}^6$ as the wrench applied at the $i$-th contact, and $\boldsymbol{F} = [F_1,\dots,F_N]^T$ as a vector of grasp forces, where $F_i$ is the grasp force at the $i$-th contact.
The minimal required grasp force to resist a target wrench $\t$ is computed by:
\begin{equation}
\begin{aligned}
& \underset{\boldsymbol{F},\{\boldw_1,\ldots,\boldw_N\}}{\text{minimize}}
& & \boldsymbol{F} \bigcdot \boldsymbol{1}_{N} \\
& \text{subject to}
& & \matr{G} \begin{bmatrix} \boldsymbol{w}_1 \\ \vdots \\ \boldsymbol{w}_N \end{bmatrix} = \t, \\
&&& \boldw_i \in \mathcal{C}_i, \hspace{0.1em} \forall{i}.
% &&& \hat{\boldsymbol{n}}_1 \bigcdot (\boldw_1 - F_1 \hspace{0.1em}  \hat{\fperp}_1) \leq  F_1 \hspace{0.1em} \hat{\boldsymbol{n}}_1  \bigcdot \hat{\boldsymbol{p}_1}. \\
% &&& \hspace{7em} \vdots \\
% &&& \hat{\boldsymbol{n}}_N \bigcdot (\boldw_N - F_N \hspace{0.1em}  \hat{\fperp}_N) \leq  F_N \hspace{0.1em} \hat{\boldsymbol{n}}_N \bigcdot \hat{\boldsymbol{p}_N}. 
\end{aligned}
\label{eq:sol_min_force}
\end{equation}

\mysum{work and quality }

Based on Hooke's law, the work $\work$ is computed by: % the displacement of the $i$-th finger $d_i$ and  of the $n$-fingered grasp is computed by:
\begin{equation}
% 	\work = \sum_{i=1}^N F_i \bigcdot d_i \text{, with } d_i = \frac{F_i}{s_{o_i}} + \min \{ \frac{F_i}{s_{c}}, d_c \},
	\work = \sum_{i=1}^N F_i \bigcdot d_i \text{, with } d_i = \frac{F_i}{s_{o_i}} + \epsilon,
\end{equation}
where $d_i$ is the displacement of the $i$-th jaw and $s_{o_i}$ is the object stiffness at contact~$i$.
$\epsilon$ is a small number, such that the minimal work quality metric is applicable to rigid objects or objects contain a rigid part. 
The displacement $d_i = \epsilon$ and the minimal work quality reduces to comparing minimal force between grasps.
%Their sum forms the finger displacement $d_i$. 
% The object stiffness is experimentally determined as described in Sec.~\ref{sec:stiffness_acquisition}.

% \mysum{compute displacement}
% To obtain the deformability at different locations of an object, we use Nvidia Flex to simulate each grasp with the force $\boldsymbol{F}_s$. 
% The simulation returns the object deformation $\boldsymbol{d}_i$ at each contact $i$.
% The deformation $d_i$ at $i$ with the grasp force $\boldsymbol{F_i}$ is computed by:
% \begin{equation}
% 	\boldsymbol{d}_i = \frac{\norm{\boldsymbol{F}_i}}{\norm{\boldsymbol{F}_s}} \boldsymbol{d}_s.
% \end{equation}

Finally, the quality $q_w$ of a grasp under $K$ perturbations of grasp pose is computed by:
\begin{equation}
    q_w = \frac{1}{K} \sum_{k=1}^K \left(1 - \frac{\work_k}{\work_{\max}}\right), 
\end{equation}
where $\work_{\max}$ is used for normalization and is the the product of maximal grasp force and object's deformation of collected data.

% Qualities $q_w$ and $q_f$ are compared in Sec.~\ref{sec:res_qualities}
% The g in rasp candidate that reqcompareduires the minimum work to complete the task has the highest score.
% $\alpha$ is a small positive constant to prevent dividing by zero.  
% The proposed quality metric is suitable for grasping rigid objects as well, since the metric reduces to the minimal force to complete the task.

%% file: sections/6-implementation.tex
\label{sec:implementation}
To compute the minimal required grasp force, classical wrench-based grasp analysis first model the possible friction and normal wrench of each contact and then estimate the total wrench that a grasp can exert on an object. 
The estimation of friction and normal wrench highly depend on the contact details.
We first describe the method used in this work to estimate contact area and pressure distribution.

\subsection{REACH: contact geometry}
Danielczuk et al.~\cite{danielczuk2019reach} proposed the Robust Efficient Area Contact Hypothesis (REACH) model for contact profile estimation between soft jaw pads and rigid objects. 
Given an object's geometry modeled as a triangular mesh, the contact area is computed as the constructive solid geometry intersection of the extruded polygon of the jaw with the object.
% The intersection provides a deformation function $\delta: \mathbb{R}^3 \rightarrow \mathbb{R}$ that models the distance that the soft pad presses into the object at each point on the contact surface. 
The intersection estimates the deformation of the soft pad around the object at each point on the contact and the pressure distribution linearly scales with the gripper pad deformation. 
The REACH model provides contact area consists of triangles and the normal pressure of each triangle. 

We apply the REACH model to estimate contact information between compliant jaw pads and deformable hollow objects due to its highly computational efficiency compare to e.g. the Finite Element Method. 
We note that the obtained contact details may not be accurate for objects with high deformability. 

\subsection{6D ellipsoidal limit surface}
Grasps with soft jaws may result in a non-planar contact area, where the friction wrench applied at the contact is six-dimensional (6D).
This work uses the 6D ellipsoid proposed in \cite{xu2019tro} as the limit surface model. % and efficient sampling and fitting strategy proposed in [other ICRA].  
Note that other friction models are applicable to the proposed work computation algorithm as well.

Friction depends on relative body motion. 
% The instantaneous velocity  of a body in 3D space consists of a rotation about an instantaneous axis of rotation (IAR) and a transnational velocity parallel to the IAR, where its magnitude is defined as the pitch.
For a given contact area and pressure distribution, possible friction wrenches are obtained by sampling the instantaneous relative motion, defined as body twist in screw theory \cite{murray.2017}. 
The direction of frictional force of each triangle is obtained by projecting the velocity onto the triangle plane, while the magnitude is the product of the coefficient of friction and the normal force applied at the triangle. 
The frictional torque is computed with respect to the friction-weighted center of pressure. 
By summing up the friction contribution of each triangle, we obtain a friction wrench of the contact for each sampled twist. 

%By using least square method [other ICRA], 

We fit the friction wrenches to a 6D ellipsoid by solving a convex optimization. %, which is described with the matrix $\matr{A} \in \mathbb{R}^{6 \times 6}$. 
The friction wrench is constrained to be in the interior of the ellipsoid.
The surface of the fitted ellipsoid is then evenly sampled to acquire linear constraints.

We use a 6D ellipsoidal limit surface to model the friction wrenches of each contact.
Given the ellipsoid matrix $\matr{A} \in \mathbb{R}^{6 \times 6}$, the friction wrenches $\f \in \mathbb{R}^6$ are constrained by:
\begin{equation}
    \f ^T \matr{A} \f \leq 1.
\end{equation}

For computational efficiency, we compute the linearized constraints $\mathcal{C}_i$ of the $i$-th contact by evenly sample the ellipsoid with $M$ points $\boldsymbol{p} \in \mathbb{R}^{6 \times M}$ on the surface of $\matr{A}$.
Each point and the outward normal of $\matr{A}$ at the point form a hyperplane. 
The friction wrenches are constrained in the interior of the $M$ hyperplanes:
\begin{equation}
\begin{aligned}
        \mathcal{C}_i = \{\boldw_i \in \mathbb{R}^6 \ | \ \boldsymbol{n} & \bigcdot \left( \boldw_i - \fperp \right)  \leq \boldsymbol{n} \bigcdot  \boldsymbol{p} \},  \\
        \text{with } \boldsymbol{n} &= \matr{A}  \boldsymbol{p}, 
\end{aligned}
\end{equation}
where $\fperp \in \mathbb{R}^6$ is the wrench impressed by the normal pressure, $\boldsymbol{n}$ are the normals of $\matr{A}$ at $\boldsymbol{p}$.

% Denote $\fperp \in \mathbb{R}^{6}$ as the wrench impressed by the normal force and denote $\boldw = \fperp + \f$ as the total wrench a contact can exert, the linear constraints for $\boldw$ are:

% We set a simplified convex optimization program to fit $N$ wrench samples compare to [TRO], where the twist samples are used for fitting in [TRO]:
% \begin{equation*}
% \begin{aligned}
% & \underset{\matr{A}}{\text{minimize}}
% & & \sum_{i=1}^N \errorwrenchi \\
% & \text{subject to}
% & & \norm{ \f_i ^T \matr{A} \f_i - 1} \leq \errorwrenchi, \forall i \in  \{1 \ldots N\}, \\
% %&&& \norm{(A^T + A)\boldw_i - \boldsymbol{s}_i \boldt_i} \leq \varepsilon_{\boldt_i}, \\
% % &&& \norm{\nabla f_1(\boldw_i,\matr{A}) - s_i \boldt_i} \leq \errortwisti \text{, with } {s}_i >0, \\
% &&& \matr{A} \succeq 0.
% \end{aligned}
% \end{equation*}

% \subsection{Construct grasp wrench space}

% \subsection{Compute minimal force based on the GWS}
% For a set of tasks t, ...
% describe Kurg's work, use LP to solve it. Requires however building the GWS.  
% They show Minkowski sum GWS is better than union because union leads to underestimation. 
% But convex hull of minkowski sum is slow. 

% the scaling factor of $q$ is essentially $\frac{1}{q}$ of GWS scaling factor. 
% min force $\lambda$ is $\frac{1}{q}$

% For an unknown task, t is modeled with a 6D unit sphere. So compute the $\epsilon$, then min force is $\frac{1}{\epsilon}$.

%% file: sections/7-experiments.tex
\subsection{Acquisition of object stiffness}
\label{sec:stiffness_acquisition}
The object stiffness is required to compute the required work of gripper jaws for a specific task. 
We estimate the object stiffness by means of its 3D mesh and physical experiments. 
One can use e.g. the Finite Element Analysis to compute the object deformation with a closing force of the gripper. 
However, the deformability of e.g. plastic bottles and cups highly depends on the wall thickness, the geometry, and material of the object, which are non-trivial to obtain.  
Therefore, we use a physical robot to collect object's stiffness at different locations in this work. 

We estimate the object's stiffness based on 1) a known gripper closing force $F_c$, 2) the gripper opening $L_s$ when its just in contact with the object, and 3) the opening $L_e$ when $F_c$ is reached. 
We first plan antipodal grasps in simulation for each object. 
$L_s$ is estimated by checking the minimal distance between the 3D object mesh and the gripper mesh at the grasp location. 

At each planned grasp location, the Robotiq 2F-85 gripper closes with a minimal possible force $F_c=20N$.
The object's stiffness $s_{o_i}$ at the location $i$, obtained by the two intersection points of the grasp axes and the object surface is computed by:
\begin{equation*}
s_{o_i} = \frac{F_c}{L_{s_i} - L_{e_i}}.
\end{equation*}
We repeat each grasp 5 times and the median of the collected gripper opening after reaching the closing force is selected as $L_e$. 

We note that the actual grasp force of the Robotiq gripper depends on the object's material and the gripper closing speed. 
Fig. \ref{fig:repeatability} shows experiment results of the repeatability of the gripper. 
An object with an open lid is grasped at five locations, where the object has different deformability at each location.
The gripper closes repeatedly with the minimal force and multiple speeds, ranges from 0 to 255, where 0 is the lowest speed. 
For each grasp location and each closing speed, the gripper closes 15 times and the gripper opening (cm) is recorded after the force is reached.
We restart the gripper program every three closes.

Fig. \ref{fig:repeatability} shows the mean and standard deviation of the gripper opening at each grasp location with a closing speed. 
Results show the the gripper opening has a higher error at grasp pose 1-3 than pose 4-5. 
This observation suggests when the gripper opening is small, e.g. when the object is thin or has large deformation, the gripper repeatability is lower than with a large gripper opening.   

Although the absolute gripper opening and the collected object's stiffness may not be highly accurate, the relative object's stiffness is reasonable, which is essential to compute optimal grasp placements. 

\input{sections/figures/robotiq_repeatability.tex}

\subsection{Simulation results: Planned grasps with three quality metrics}

We plan grasps on five objects based on their 3D meshes and the interpolated stiffness maps, as shown in Fig. \ref{fig:metrics}(a) and (b), where two objects are 3D printed using NinjaFlex TPU flexible filament since we want to experiment with objects of various shapes and materials. 
% The stiffness map shows the stiffness at different grasp locations and the interpolated stiffness 
Note that some inaccuracies of the measured stiffness exist, such as the rigid cap of the object~1 and the neck of object~2.

We sample antipodal grasps candidates and compute grasp quality for 1) vertical lifting and 2) lifting and $90^\circ$ rotation task.
Three quality metrics are compared: grasp reliability, minimal force, and minimal work, as shown in Fig. \ref{fig:metrics}. 
The two tasks are modeled with a 6D gravity wrench to be resisted under one and three object poses obtained by discretizing the manipulation trajectory, respectively, since the gravity wrench remains the same for the vertical lifting task. 
The metric of each grasp is computed under 20 perturbations of grasp poses for each object pose of the task. 
The lowest quality value of a grasp among all object poses is selected as the value for each metric.

% where the stiffness of each contact location uses the nearest point in the interpolated stiffness map. 
% \subsubsection{Planned grasps with three quality metrics}
% \label{sec:res_qualities}
% We show the quality of grasp candidates under perturbations in grasp pose for five objects. 
% The target wrench of each task is modeled as a single gravity 6D wrench. 
% The rotation task is modeled as 3 poses along the trajectory and each grasp computes grasp score for object pose.
% The lowest score along all poses is the quality of the grasp.  
Fig. \ref{fig:metrics} suggests the planned grasps with the proposed minimal work quality metric avoid causing large deformations of the object, while are able to resist the gravitational disturbances of the manipulation tasks. 
%Results show false positives can be largely reduced with the minimal work metric, as fewer grasp candidates are considered to be successful.

% \input{sections/figures/grasp_scores.tex}
% \input{sections/figures/grasp_score_rot1.tex}
% \input{sections/figures/failed_grasps.tex}

\input{sections/figures/quality_metrics.tex}
\input{sections/figures/fig_phys_exp.tex}

\input{sections/figures/tab_phys_exp.tex}
\subsection{Physical experiments}
We evaluate the planned grasps for three representative objects (object 1,3,5) with physical experiments for the two manipulation tasks.
We select 46 grasp poses in total for three objects that cover different regions of object and each grasp is repeated five times for each task, which sums up to 460 grasps in total.
We consider a grasp is successful if the task is completed while the object is returned into its original shape when the grasp force is released and the content is not dislodged during the grasp. 

Objects are filled with wet towels to imitate the weight filled with liquid without changing the object's deformability or damage the electrical devices. 
Object 1 and 3 are sealed with a balloon to infer the content spillage.
By measuring the balloon's inflation amount before and after the grasp, the content is considered spilled if the inflation difference is larger than a threshold. 

We use the balanced accuracy score to evaluate the prediction accuracy, which is the raw accuracy balanced by the number of successful and failed grasps in the collected data.
Table \ref{tab:phys_res} shows the balanced accuracy of the three grasp quality metrics for the two manipulation tasks and Fig. \ref{fig:phys_exp} shows examples of correct and wrong prediction of the planned minimal work grasps. 
The proposed metric reaches 74.2\% and 71.3\% balanced accuracy for the two tasks. respectively, and are up to 24.2\% higher than the grasp reliability and the minimal force metric.
However, we note that the prediction accuracy is relatively low for object 5 (a plastic cup) compared to other objects.
% One reason is that the number of succeeded grasps for the plastic cup is low.
% This leads to a low AP value if some true positives are not correctly predicted. 
% Apart from that, wrong predictions exist. 
% We note that false positives exist e.g. when the object is grasped in the middle and rotated. 
This suggests the proposed algorithm for minimal work computation may not perform well for objects with large deformations.
Furthermore, the possible minimum grasp force with the Robotiq Gripper is much larger than the planned force and causes large object deformations and false positives. 
% False negatives are observed when grasping the top or the bottom to rotate the cup. 

% real object, stiffness map, robust, min force, min work

% We evaluate the minimal work grasp quality for different tasks with the target wrench $\boldsymbol{t}$ and for a unknown task.
% The tasks include:
% \begin{itemize}
% 	\item Lifting vertically
% 	\begin{itemize}
% 		\item $\boldsymbol{t} = [0,0,mg,0,0,0]^T$
% 	\end{itemize}
% 	\item Transportation along the $x$-axis (not included yet)
% 	\begin{itemize}
% 		\item $\boldsymbol{t} = [ma,0,mg,0,0,0]^T$
% 	\end{itemize}
% 	\item Transportation along the $y$-axis (not included yet)
% 	\begin{itemize}
% 		\item $\boldsymbol{t} = [0,ma,mg,0,0,0]^T$
% 	\end{itemize}
% 	\item pouring around the $x$-axis
% 	\begin{itemize}
% 		\item $\boldsymbol{t} = [0,0,mg,mg\bigcdot l,0,0]^T$
% 	\end{itemize}
% 	\item pouring around the $y$-axis
% 	\begin{itemize}
% 		\item $\boldsymbol{t} = [0,0,mg,0,mg\bigcdot l,0]^T$
% 	\end{itemize}
% 	\item pouring around $x,y,z$-axes
% 	\begin{itemize}
% 		\item $\boldsymbol{t} = [0,0,mg,\tau_x,\tau_y,\tau_z]^T$
% 	\end{itemize}
% \end{itemize}  

%% file: sections/figures/robotiq_repeatability.tex
\begin{figure}
	\vspace{0.5em}

	\centering	
\includegraphics[width= 90 mm,trim={0cm  22.2cm 2cm 0cm},clip]{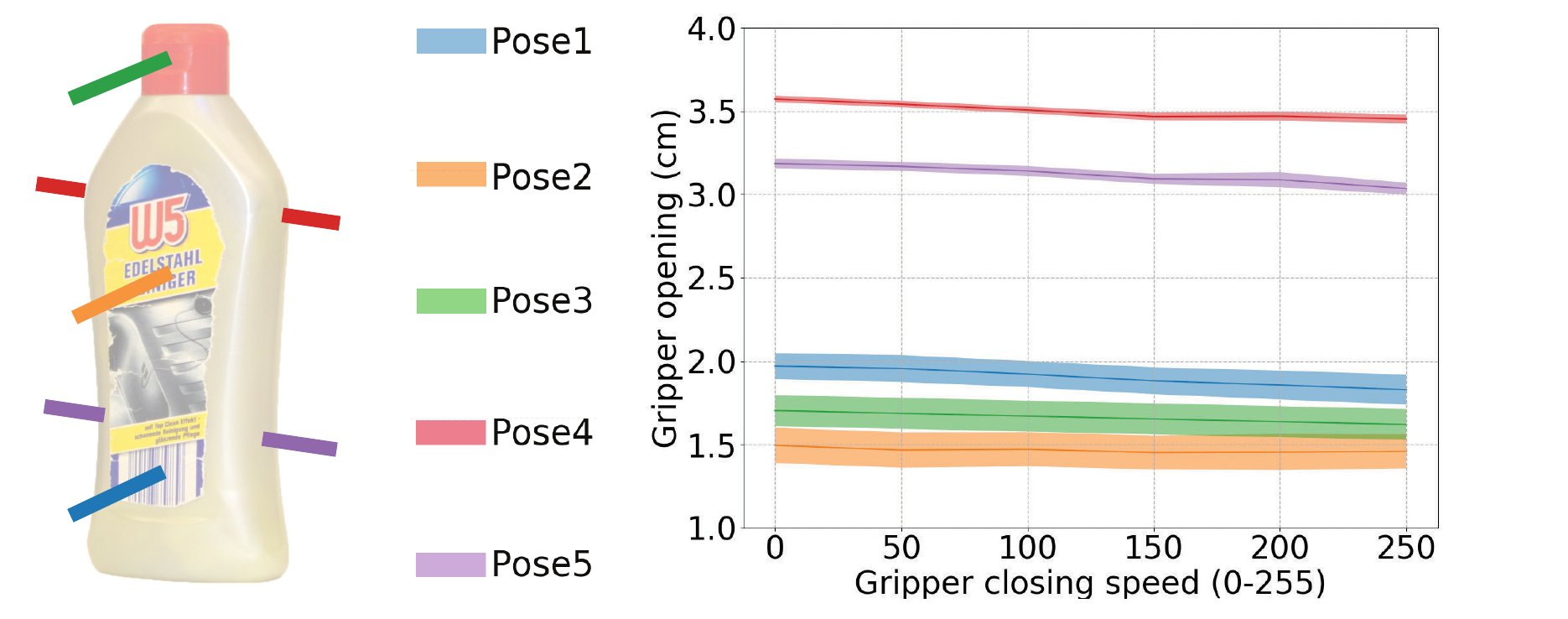}
	\caption{Robotiq gripper repeatability test.}
	\label{fig:repeatability}
\end{figure}

%% file: sections/figures/quality_metrics.tex
\begin{figure*}
	\centering	
	\vspace{0.5em}
\includegraphics[width=175 mm,trim={0cm  11cm 0cm 0cm},clip]{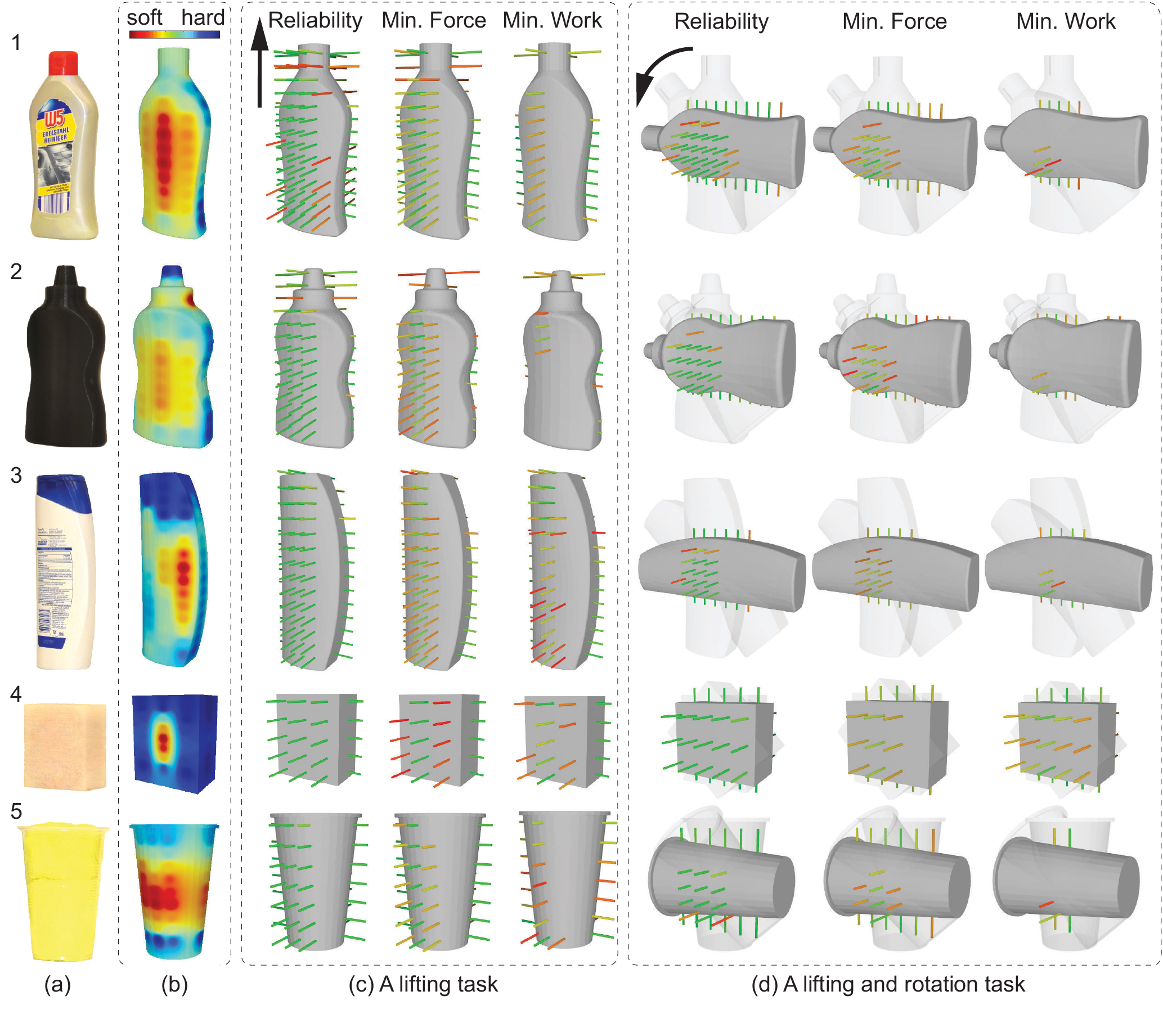}	
	\caption{Planned grasps for five physical objects. Object 2 and 4 are 3D printed using NinjaFlex TPU flexible filament. (b): The interpolated stiffness map. Grasp qualities with three metrics for a lifting task, and (c) a lifting and $90^\circ$ rotation task (d). The proposed minimal work metric computes grasps that resist gravitational disturbances without causing large deformation.}
	\label{fig:metrics}
\end{figure*}

%% file: sections/figures/fig_phys_exp.tex
\begin{figure}
	\centering	
		\vspace{0.5em}

\includegraphics[width=85 mm,trim={0cm  0cm 0cm 0cm},clip]{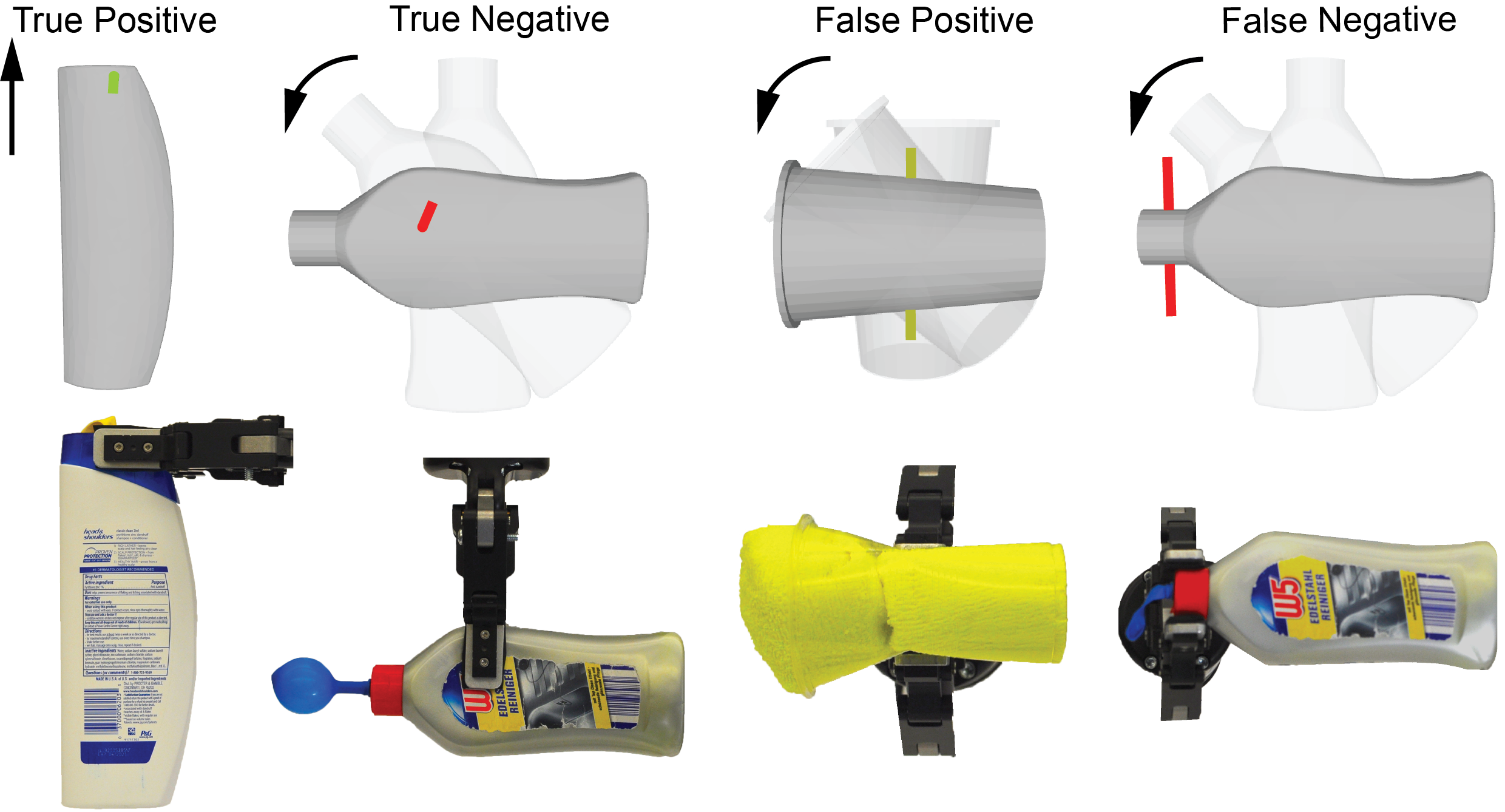}
	\caption{Examples of grasp success predictions with the minimal work quality metric. A grasp is considered successful if the manipulation task succeeded while the content is not dislodged. An inflated balloon suggests liquids in the container might dislodge.}
	\label{fig:phys_exp}
\end{figure}

%% file: sections/figures/tab_phys_exp.tex
% \begin{table}
% 	\begin{center}
% 		\caption{Mean average precision of three quality metrics: grasp reliability (GR), minimal force (MF), and the proposed minimal work (MW). }
% 		\begin{tabular}{|c|c|c|c|c|c|c|}
% 			\hline
% 			\multirow{2}{*}{Object} & \multicolumn{3}{c|}{Lifting} &  \multicolumn{3}{c|}{Lifting and $90^\circ$ rotation} \\ \cline{2-7}
% 			& GR & 
% 			MF & MW & GR & MF & MW \\ \hline
% 	        1 & 0.282 & 0.594& \textbf{0.832}& 0.353& 0.752 & \textbf{0.821} \\ \hline
% 	        3 & 0.442 & 0.555 & \textbf{0.845}& 0.313 & 0.467 & \textbf{0.715}\\ \hline
% 	        5 & 0.602 & 0.566& \textbf{0.856} & \textbf{0.264}& 0.172& 0.199\\ \hline
% 	        \hline
% 	        All & 0.442 & 0.571& \textbf{0.844}& 0.310 & 0.464& \textbf{0.578 }\\ \hline
% 		\end{tabular}
% 	\label{tab:phys_res}
% 	\end{center}
% \end{table}

\begin{table}
	\begin{center}
		\caption{Balanced accuracy of three quality metrics: grasp reliability (GR), minimal force (MF), and the proposed minimal work (MW). }
		\begin{tabular}{|c|c|c|c|c|c|c|}
			\hline
			\multirow{2}{*}{Object} & \multicolumn{3}{c|}{Vertical lifting} &  \multicolumn{3}{c|}{Lifting and $90^\circ$ rotation} \\ \cline{2-7}
			& GR & 
			MF & MW & GR & MF & MW \\ \hline
	        1 & 0.500 & 0.767 & \textbf{0.833}& 0.575& 0.785 & \textbf{0.875} \\ \hline
	        3 & 0.500 & 0.687 & \textbf{0.851}& 0.516 & 0.714 & \textbf{0.813}\\ \hline
	        5 & 0.500 & 0.392& \textbf{0.542} & \textbf{0.680} & 0.630& 0.450 \\ \hline
	        \hline
	        All & 0.500 & 0.615 & \textbf{0.742} & 0.590 & 0.710 & \textbf{0.713 }\\ \hline
		\end{tabular}
	\label{tab:phys_res}
	\end{center}
\end{table}

%% file: sections/8-discussion.tex
We propose a minimal work quality metric to plan grasps for 3D deformable hollow objects. 
% We first compute the minimal required force to resist a target wrench by solving a linear program.
We evaluate the proposed metric with real-world grasps for a vertical lifting and for a lifting and $90^\circ$ rotation task. 
Physical experiments suggest that 74.2\% and 71.3\% balanced accuracy can be achieved for the two tasks, respectively, and up to 24.2\% higher than the classical wrench-based quality metrics. 
% We apply the REACH model in this work since the gripper pads are much softer than the hollow objects.
% We think REACH provides a reasonable approximation and is highly computational efficient compared to e.g. an FEM simulation. 
\subsection{Limitations}
We note that the proposed method may not perform well for objects that having large deformations. 
An Finite Element Method can be used to compute the deformation of such objects and the pressure distribution of the contact. 
The grasp quality can then be computed based on objects's deformed shape.
Furthermore, one reason that leads to false positives is that the actual possible minimal grasp force of the Robotiq gripper is higher than specified in the planned grasps. 
We plan to use e.g. a Schunk gripper mounted with force sensors to plan grasps with computed minimal grasp force.
% Furthermore, object deformation does not necessarily lead to a smaller volume and the content will not    spl, .g. grasping on the side of a plastic bottle.

% \subsubsection{Object stiffness}
% The measured object stiffness is not accurate because of the repeatability of the robotiq gripper. 

% future work 
\subsection{Future work}

We note that the minimal work grasp is applicable to grasping rigid objects, when the jaw pads are deformable.
We intend to further investigate this duality and plan grasps for both rigid and deformable objects.

%% file: sections/10-ackowledgement.tex
\section{Acknowledgments}
\footnotesize
This research was performed at the AUTOLAB at UC Berkeley in affiliation with the Berkeley AI Research (BAIR) Lab, Berkeley Deep Drive (BDD), the Real-Time Intelligent Secure Execution (RISE) Lab, and the CITRIS ``People and Robots" (CPAR) Initiative. The authors were supported in part by the Scalable Collaborative Human-Robot Learning (SCHooL) Project, NSF National Robotics Initiative Award 1734633 and by donations from Google, Siemens, Amazon Robotics, Toyota Research Institute, Autodesk, ABB, Samsung, Knapp, Loccioni, Honda, Intel, Comcast, Cisco, Hewlett-Packard and by equipment grants from PhotoNeo and NVIDIA. Any opinions, findings, and conclusions or recommendations expressed in this material are those of the author(s) and do not necessarily reflect the views of the Sponsors. We thank our colleagues who helped with experiments and provided helpful feedback and suggestions, especially William Wong, Priya Sundaresan, Harry Zhang, Jackson Chui, Kate Sanders, and Andrew Lee. 